\documentclass[letterpaper]{article} 
\usepackage{aaai23}  
\usepackage{times}  
\usepackage{helvet}  
\usepackage{courier}  
\usepackage[hyphens]{url}  
\usepackage{hyperref} 
\usepackage{graphicx} 
\usepackage{natbib}  
\usepackage{caption} 
\usepackage{algorithm}
\usepackage{algorithmic}

\usepackage{subfigure}
\usepackage{bm}
\usepackage{amsmath}
\usepackage{amssymb}
\usepackage{todonotes}
\usepackage{cleveref}
\usepackage{paralist}
\usepackage{lipsum}
\usepackage{diagbox}
\usepackage{newfloat}
\usepackage{listings}
\DeclareCaptionStyle{ruled}{labelfont=normalfont,labelsep=colon,strut=off} 
\floatstyle{ruled}
\newfloat{listing}{tb}{lst}{}
\floatname{listing}{Listing}

\DeclareMathOperator {\mlp}{MLP}
\DeclareMathOperator {\lstm}{LSTM}
\DeclareMathOperator {\kge}{KGE}
\DeclareMathOperator {\emb}{emb}
\DeclareMathOperator{\HuberLoss}{HuberLoss}

\title{A Machine with Short-Term, Episodic, and Semantic Memory Systems}
\author {
    Taewoon Kim\textsuperscript{\rm 1},
    Michael Cochez\textsuperscript{\rm 1},
    Vincent François-Lavet\textsuperscript{\rm 1},
    Mark Neerincx\textsuperscript{\rm 2},
    Piek Vossen\textsuperscript{\rm 1}
}
\affiliations {
    \textsuperscript{\rm 1}Vrije Universiteit Amsterdam\\ \textsuperscript{\rm 2}Technische Universiteit Delft\\
    \{t.kim, m.cochez, vincent.francoislavet, p.t.j.m.vossen\}@vu.nl,
    m.a.neerincx@tudelft.nl
}

\begin{document}

\maketitle

\begin{abstract}
Inspired by the cognitive science theory of the explicit human memory systems, we have modeled an agent with short-term, episodic, and semantic memory systems, each of which is modeled with a knowledge graph.
To evaluate this system and analyze the behavior of this agent, we designed and released our own reinforcement learning agent environment, ``the Room'', where an agent has to learn how to encode, store, and retrieve memories to maximize its return by answering questions.
We show that our deep Q-learning based agent successfully learns whether a short-term memory should be forgotten, or rather be stored in the episodic or semantic memory systems.
Our experiments indicate that an agent with human-like memory systems can outperform an agent without this memory structure in the environment. \textbf{The code is open-sourced at \href{https://github.com/humemai/agent-room-env-v1}{https://github.com/humemai/explicit-memory}.}
\end{abstract}

\section{Introduction}
\label{sec:intro}

The cognitive science theory suggests that humans have an explicit memory system, which is composed of semantic and episodic memory systems~\cite{Tulving1983-TULEOE, tulving1985memory, Tulving1973EncodingSA}. Semantic memory deals with general world knowledge, while episodic deals with one's personal memory. 
For example, when asked: ``Where are laptops \emph{usually} located?'', you might be able to answer: ``On a desk'', if you have successfully encoded and stored a relevant memory. 
However, you are unlikely able to recall when and where such memory (or knowledge) got to be stored in your memory system. 
Nonetheless, you were able to retrieve it. 
This is because this type of memory is semantic. 
Semantic memory contains (general) factual knowledge without momentary or time-bound information (i.e., where and when the event occurred). 
Let's consider another question: ``Where is Alice's laptop?'' and let us assume that you have observed where Alice's laptop was. 
To answer this question, one revisits when and where this memory was encoded and stored. 
Retrieval of such a memory is the reconstruction process of the event. 
This type of memory is called episodic. 
These are more individual to you than semantic memories, include information about personal experiences, and often have an affective connotation.

Nowadays, machines have become reasonably good at answering factual knowledge questions. 
Search engines and virtual assistants can retrieve relevant information (often harvested from the Internet) 
and answer factual and commonsense questions (i.e., those probing for semantic knowledge). 
However, these systems will not remember what they did yesterday (i.e., episodic memory). 
This is because they were not designed to answer such questions.

Consider, for example, a robot deployed to a nursery home whose job is to remind the elderly to take their pills every day. 
If such robot can successfully remember when and where it has seen them taking the pills, by observing the environment, it is a great help to them. 
However, it is not feasible for an agent to encode and store every observation as an episodic memory, as memory capacity is bounded. 
If the information is no longer stored, the agent can use commonsense knowledge (semantic memory) and say ``You left your pills in the cabinet''.

Motivated by this type of examples, we have modeled an agent that has both semantic and episodic memory systems. 
The agent interacts with the environment and answers questions to maximize its reward. 
To make it more realistic, our agent can only partially observe the environment and hence it needs a memory system to be able to answer. 
Our hypothesis is that if it has two explicit memory systems, rather than one, it can answer the questions more successfully.

Knowledge graphs (KG) model data in the form of a graph, where entities are linked with directed relations.
This way of representing data is not only readable / writable by both humans and machines, but also deductive and inductive reasoning can be performed. 
Moreover, once we convert them into knowledge graph embeddings (KGE), we can take advantage of neural networks to solve various problems~\cite{Hogan_2022}. 

Although KGs can be built by experts with their domain knowledge, they are more powerful when they are open and crowdsourced, since they can take advantage of collective human knowledge.
Wikidata~\cite{10.1145/2629489} and ConceptNet~\cite{Speer_Chin_Havasi_2017} are some of the examples. 
ConceptNet tries to capture the meanings of the words, resulting in capturing common sense knowledge (e.g., \texttt{(laptop, AtLocation, desk)}). 
We used part of their knowledge graph for our experiments.

Reinforcement learning (RL) has become increasingly popular over the past few years~\cite{rl_survey}. Deep Q-learning~\cite{DBLP:journals/corr/MnihKSGAWR13}, an RL that uses deep learning, has showed us that an RL agent can learn how to play Atari games. It finds an optimal policy by finding the optimal action-value function. RL is especially a powerful method, when it is infeasible for us humans to label every best state-action pair that the agent can take.

The contributions of this paper are as follows.
\begin{inparaenum}[1)]
	\item Inspired by the cognitive science theory, we explicitly model an agent with human-like memory systems (i.e., short-term, episodic, and semantic), each of which is modeled with a knowledge graph. 
	\item We designed and released our own environment, compatible with OpenAI Gym~\cite{brockman2016openai}, where an agent has to learn how to encode, store, and retrieve memories to maximize its return by answering questions. 
	\item Our deep Q-learning based RL agent has successfully learned a memory management policy by interacting with the environment.
\end{inparaenum}

The rest of this paper is organized as follows. 
In Section~\ref{sec:methodology}, we propose our method, which allows an agent to learn how to encode observations, and to store and retrieve memories. 
In Section~\ref{sec:experimentalsetup}, we outline how we carried out our experiments and show the results in Section~\ref{sec:results}. In Section~\ref{sec:related}, we compare our work with other works, and show how ours differs from theirs. 
Finally, in Section~\ref{sec:conclusions}, we conclude this paper and mention some possible future research directions.

\section{Methodology}
\label{sec:methodology}

\begin{figure}[tb]
\centering
\includegraphics[width=0.5\columnwidth]{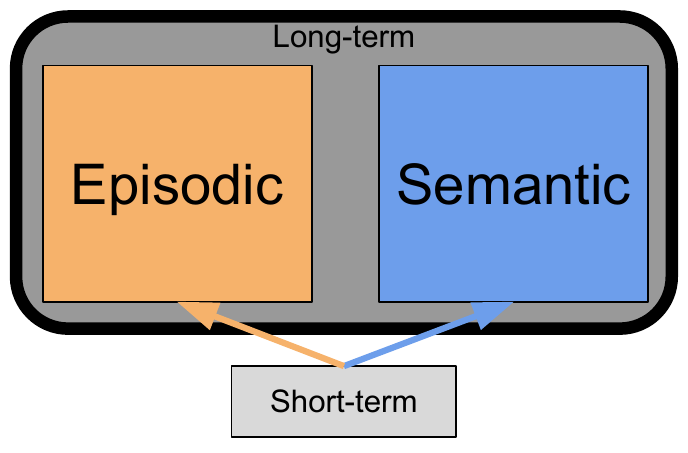}
\caption{The memory systems of the agent. 
The long-term (explicit) memory systems consist of episodic and semantic memory systems.}
\label{fig:memory_systems}
\end{figure}

To perform our experiments, we have to design an agent which mimics the memory systems.
Besides, we need an environment in which the agent operates, as well as define how the agent observes the environment.

\subsection{The Room Environment}
\label{sec:roomenv}

We model our environment with a discrete-event simulation (DES). 
In this simulation, there is a room with $N_{humans}$ humans, $N_{objects}$ different types 
of objects, and $N_{object\_locations}$ possible object locations. 
At each time step, every human can place an object at one of the possible object locations, within their capacity limit. 
For example, if the capacity of a desk is four, then only four objects can be placed on the desk at once. 
Each change in the room state is called an event. For example, let us assume that Alice was the only person in the room. If at $timestep=1$, Alice's laptop was on the desk, but at $timestep=2$, it is on her lap, then the event at $timestep=2$ is the change of the state.

The humans in the simulation do not move other humans' objects, but only theirs. They also have their own personal fixed routine. 
For example, Alice has her laptop on the desk two days in a row, and then the next three days she has it on her lap. She repeats this routine. 
The actual routines of the humans in our experiments are longer and involve more objects and object locations than this example.

The object locations on which each human places their object is not uniformly random. 
With a given probability $p_{commonsense}$, they place their objects on a commonsense location (e.g., a bowl on the cupboard), and with $1 - p_{commonsense}$ chance, they place them on a non-commonsense location (e.g., a bowl in the wardrobe). 
As ConceptNet~\cite{Speer_Chin_Havasi_2017} can have multiple commonsense object locations per object, we chose \emph{commonsense location} of an object as the one with the highest weight.

We make our environment compatible with OpenAI-Gym by creating a wrapper around the DES, where an agent only partially observes the entire state. The agent goes around the room and observes one human at a time (e.g., At $timestep=42$, the agent observes that Alice has her laptop on her lap).
After each observation, the agent is asked a question about the current state of the DES (e.g., Where is Alice's laptop?).
If it answers correctly, it gets a reward of +1 and if not, it gets a reward of 0. 

The observations made are given as quadruples, $(h, r, t, timestamp)$.
Here $h$, $r$, and $t$ stand for head, relation, and tail, respectively.
This means that the agent observes that the entity $h$ has the relation $r$ to the entity $t$ at the given timestamp (e.g., \texttt{(Bob's laptop, AtLocation, desk, 42)}).
The question is given as a triple $(h, r, ?)$, such as \texttt{(Bob's laptop, AtLocation, ?)}. 

\subsection{Human-Like Memory Systems}
\label{sec:memory_systems}

\begin{figure}[tb]                                                                        
\centering                                                                               
\subfigure[An example of an episodic memory.]{                                                       
\includegraphics[width=0.46\columnwidth]{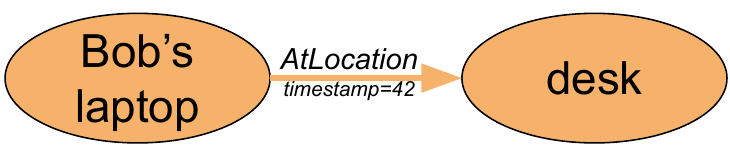}                     
\label{fig:episodic_example}                                                             
}                                                                                        
\hfill                                                                                   
\subfigure[An example of a semantic memory.]{                                                      
\includegraphics[width=0.46\columnwidth]{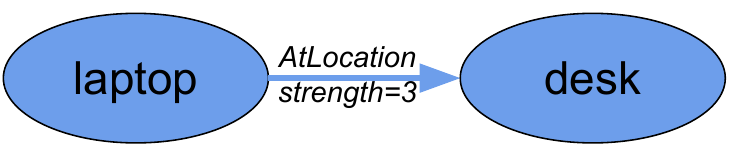}                              
\label{fig:semantic_example}                                                            
}                                                                                        
\caption{An episodic memory and a semantic memory represented as a knowledge graph.}
\label{fig:explicit_example}                                                               
\end{figure}

We model our agent with both a short-term and a long-term memory system. 
The long-term memory is split in two parts, the episodic and semantic memory.
Every observation is initially stored in the short-term memory system.
When that gets full, the agent must decide what to do with the oldest observation in short-term memory. 
It can take one of the three actions: 
\begin{inparaenum}[1)]
\item \emph{forget }it completely, 
\item move it \emph{to the episodic part} of the long-term memory system, or
\item move it \emph{to the semantic part} of the long-term memory system.
\end{inparaenum}
A diagram of this memory structure can be found in Figure~\ref{fig:memory_systems}.

There are constraints on what type of information can be stored in each memory system; we closely follow the cognitive science theories behind them~\cite{Tulving1983-TULEOE, tulving1985memory, Tulving1973EncodingSA}.
Both the short-term and episodic memory have the same quadruple format as the observation.

The facts in the semantic memory are also quadruples $(h, r, t, strength)$.
Here, $h$, $r$, and $t$ are the same as above, but rather than storing the timestamp, it has a $strength$ term, which indicates how strong this memory is. 
An example could be \texttt{(laptop, AtLocation, desk, 3)}, where 3 stands for the number of times that the agent has decided to store this memory in its semantic memory system. 
This memory system has to do with the general world knowledge, and thus $h$ is not individual-specific (e.g., \texttt{laptop} instead of \texttt{Bob's laptop}).
Figure~\ref{fig:explicit_example} shows an example of memories that could be stored.

\subsubsection{Memory Storage}
\label{sec:memory_storage}

\begin{algorithm}[tb]
	\caption{Memory retrieval}
	\label{alg:memory_retireval}
	\textbf{Input}: Question $\bm{q}$, episodic memory system $\bm{M_e}$, and semantic memory system $\bm{M_s}$.\\
	\textbf{Output}: Relevant memory $\bm{m}$.
	\begin{algorithmic}[1] 
		\IF { $\exists \bm{m_e} \in \bm{M_e}: \bm{m_e}$ is relevant for $\bm{q}$, $\nexists \bm{m_{\hat{e}}} \in \bm{M_e}: \bm{m_{\hat{e}}}$ is relevant for $\bm{q}$, $\bm{m_{\hat{e}}}$ is more recent than $\bm{m_{e}}$}
		\STATE $\bm{m} = \bm{m_e}$
		\ELSIF { $\exists \bm{m_s} \in \bm{M_s}: \bm{m_s}$ is relevant for $\bm{q}$, $\nexists \bm{m_{\hat{s}}} \in \bm{M_s}: \bm{m_{\hat{s}}}$ is relevant for $\bm{q}$, $\bm{m_{\hat{s}}}$ is stronger than $\bm{m_{s}}$}
		\STATE $\bm{m} = \bm{m_s}$
		\ELSE
		\STATE $\bm{m} = null$
		\ENDIF
		\STATE \textbf{return} $\bm{m}$
	\end{algorithmic}
\end{algorithm}

All three memory systems (i.e., short-term, episodic, and semantic) are bounded in size. 
Each of them is a knowledge graph where the number of quadruples is limited. 
We use the cognitive science theory to manage the episodic and semantic memory systems. 
That is, if the episodic memory system is full, we ``forget'' the oldest memory stored. 
If the semantic memory system is full, we ``forget'' the weakest memory. 

Dealing with the short-term memory system when it is full is more interesting and challenging.
Although the cognitive science theory suggests that the memories in this system will be either moved to the long-term memory systems or forgotten, there is no clear explanation on how this choice is made. 
Besides, this gets complicated by the ever-changing environment the agent is in. 
Therefore, we decided to learn the best behavior.
We use a deep Q-learning based agent that learns to do this by maximizing its return~\cite{DBLP:journals/corr/MnihKSGAWR13}.

\subsubsection{Memory Retrieval}
\label{sec:memory_retrieval}

In our scenario, each time an agent is asked a question, a relevant memory (i.e., quadruple) has to be retrieved from the knowledge graph to answer the question. 
For example, when the question \texttt{(Bob's laptop, AtLocation, ?)} is asked at $timestep=44$, the episodic memory \texttt{(Bob's laptop, AtLocation, desk, 42)} might be relevant. 
However, when the agent has a more recent episodic memory \texttt{(Bob's laptop, AtLocation, cupboard, 43)}, it is likely that this memory provides a more accurate answer to the question. 
Therefore, when there is more than one relevant memory in the episodic memory system, we'll have the agent choose the most recent one.

The episodic memory system alone might not contain the answer to every question. 
For example, if there are no relevant memories in the episodic memory system to answer the above question, the agent can look up its semantic memory system where it might find something like \texttt{(laptop, AtLocation, cupboard, 2)}. 
Using this general world knowledge, the agent might be able to answer the question. 
However, there might also be another relevant memory in the system such as \texttt{(laptop, AtLocation, desk, 3)}. 
Since this memory is stronger than the other, it is likely that this memory provides the correct answer to the question.
These rules form an algorithm for memory retrieval with pseudocode presented in Algorithm~\ref{alg:memory_retireval}.

\begin{algorithm}[tb]
\caption{Converting a knowledge-graph-based memory system into a learnable knowledge graph embedding (KGE).
$\mathbin\Vert$ is the concatenation operation.
}
\label{alg:kg_conversion}
\textbf{Input}: A memory system $\bm{M}$, the embedding function $\emb$ (implemented as a lookup table) \\
\textbf{Output}: The embedding of the memory system $\kge(\bm{M})$
\begin{algorithmic}[1] 
\STATE $\bm{M_{\text{sorted}}} \leftarrow$ memory system $\bm{M}$ sorted by strength (respectively timestamp)
\STATE $\kge(\bm{M}) \leftarrow$ [], i.e., an empty list

\FORALL{$(h, r, t, value) \in \bm{M_{\text{sorted}}}$ }
\IF {$\bm{M}$ is semantic memory} 
\STATE Append $\emb(h) \mathbin\Vert \emb(r) \mathbin\Vert \emb(t)$ to $\kge(\bm{M})$
\ELSIF {$\bm{M}$ is short-term or episodic memory}
\STATE $\{h_{n}, h_{o}\}$ $\leftarrow$ split $h$ into the human's name and the object
\STATE Append $\left(\emb(h_{n}) + \emb(h_{o})\right) \mathbin\Vert \emb(r) \mathbin\Vert \emb(t)$ to $\kge(\bm{M})$
\ENDIF
\ENDFOR
\STATE \textbf{return} $\kge(\bm{M})$
\end{algorithmic}
\end{algorithm}

\begin{figure}[tb]
\centering
\includegraphics[width=0.99\columnwidth]{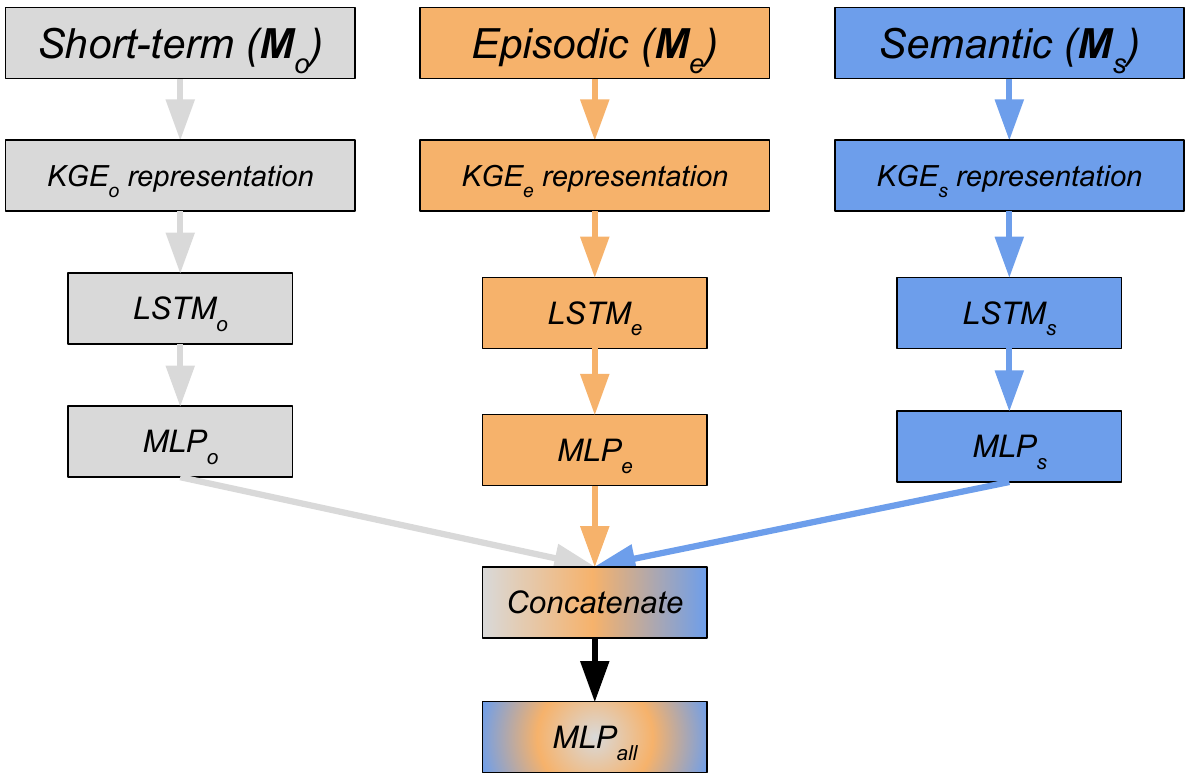}
\caption{The Q-network diagram, where the short-term $\bm{M}_{o}$, episodic $\bm{M}_{e}$, and semantic $\bm{M}_{s}$ memory systems are given as the initial input. 
The LSTM blocks output the last hidden states of the LSTMs.
The final output is the state-action values of the $Q$ function.
}
\label{fig:q_network_diagram}
\end{figure}

\subsection{The Deep Q-learning Reinforcement Learning Agent}
\label{sec:deep_q_learning_agent}

As mentioned in Section~\ref{sec:memory_storage}, we learn the task of choosing what to do with the oldest short-term memory with an RL approach. 
That is, the memory storage policy is learned by RL, while the memory retrieval policy is not learned but done by Algorithm~\ref{alg:memory_retireval}.
We have chosen deep Q-learning as our RL algorithm, because of its proven track record on other problems.

We define the state $\bm{s}$ as $(\bm{M}_{o}, \bm{M}_{e}, \bm{M}_{s})$, where the elements stand for the short-term, episodic, and semantic memory systems, respectively. 
The size of the discrete action space $|\mathcal{A}|$ is three: forget the oldest short-term memory in $\bm{M}_{o}$ completely, move it to the episodic memory $\bm{M}_{e}$, or move it to the semantic memory $\bm{M}_{s}$. Every time the agent takes an action, it also answers a given question, and a corresponding reward is given immediately. 
This reward is 1 if the answer is correct, 0 otherwise.

\begin{figure}[tb]
\centering
\includegraphics[width=0.7\columnwidth]{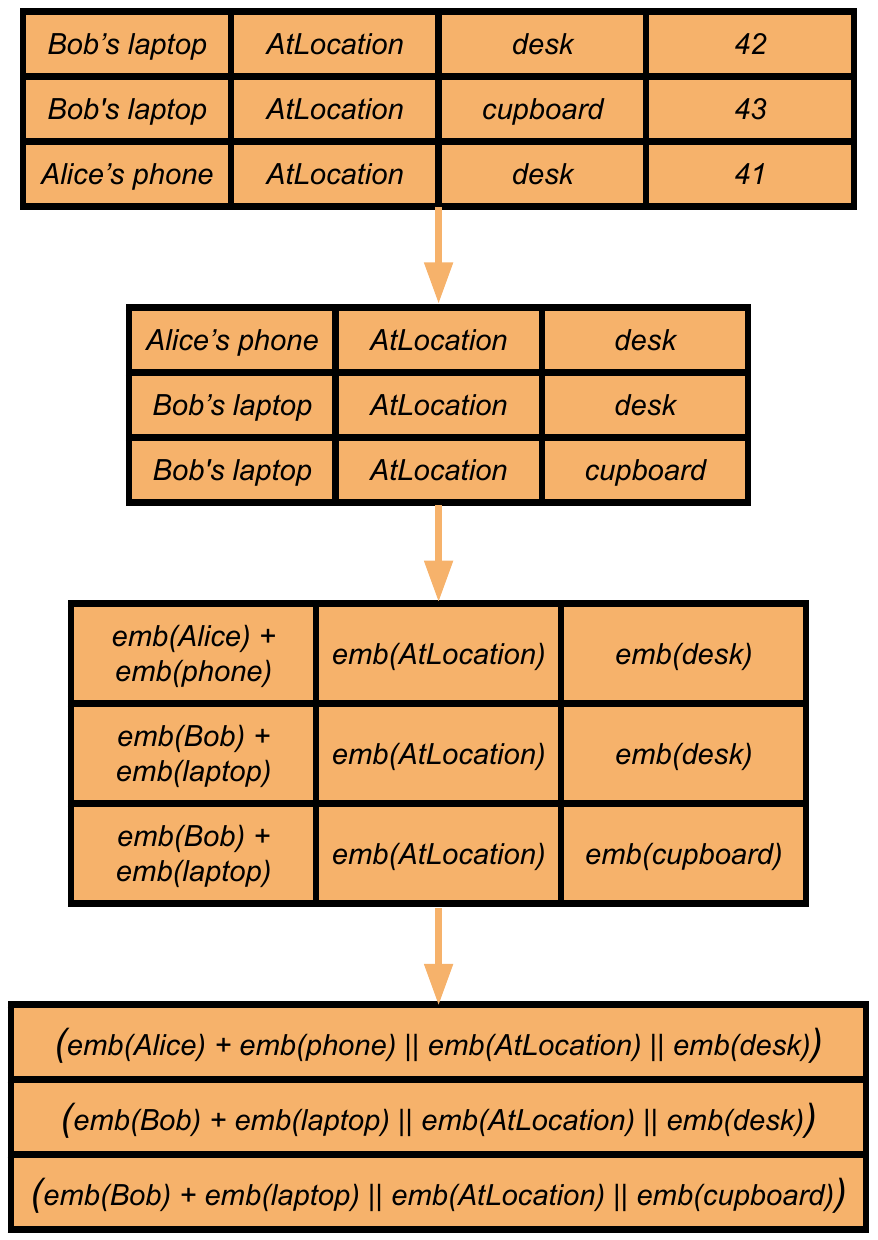} 
\caption{An example of converting an episodic memory system into a KGE representation.}
\label{fig:kge_conversion}
\end{figure}

For every entity and relation in each of the memory systems (i.e., knowledge graphs), we maintain a learnable embedding vector in the lookup table $\emb$. 
Now, in order to use a memory system as an input for the Q-network, we need to create an embedding (i.e., a numeric representation) for the complete memory system itself.
We do this by ordering the quadruples of the memory system by strength for semantic memories and timestamp for short-term and episodic. 
The embedding of the memory system is then the sequence of embeddings of quadruple in the same order.
To get the embedding of a quadruple, we retrieve and concatenate the embeddings of its head, relation, and tail.
Algorithm~\ref{alg:kg_conversion} outlines the procedure. 
Note that for the short-term and episodic, we also compute embeddings for the human actors, so that the agent can learn their individual characteristics. 
Figure~\ref{fig:kge_conversion} gives an example.

There are many sophisticated neural networks.
For our Q-network we use an LSTM-based~\cite{HochSchm97} neural network because this has shown to work well in many cases.
The Q-network is written out in Equation~\ref{eq:q_network} and the diagram of it is visualized in Figure~\ref{fig:q_network_diagram}. The output dimension of $\mlp_{all}$ is three, and each of them represents the state-action value (Q-value) of the $Q$ function. 

\begin{equation}
\begin{split}
&\text{Q-network} =\\
&\mlp_{all}\Biggl(\mlp_{o}\biggl(\lstm_{o}\Bigl(\kge_{o}\bigl(\bm{M}_{o}\bigl)\Bigl)\biggl) \mathbin\Vert\\
&\mlp_{e}\biggl(\lstm_{e}\Bigl(\kge_{e}\bigl(\bm{M}_{e}\bigl)\Bigl)\biggl) \mathbin\Vert\\
&\mlp_{s}\biggl(\lstm_{s}\Bigl(\kge_{s}\bigl(\bm{M}_{s}\bigl)\Bigl)\biggl)\Biggl)
\end{split}
\label{eq:q_network}
\end{equation}
where $\mathbin\Vert$ denotes the vector concatenation operator.

\begin{figure*}[tb]
    \centering
    \subfigure[Average loss, training.]{
    \includegraphics[width=0.3\textwidth]{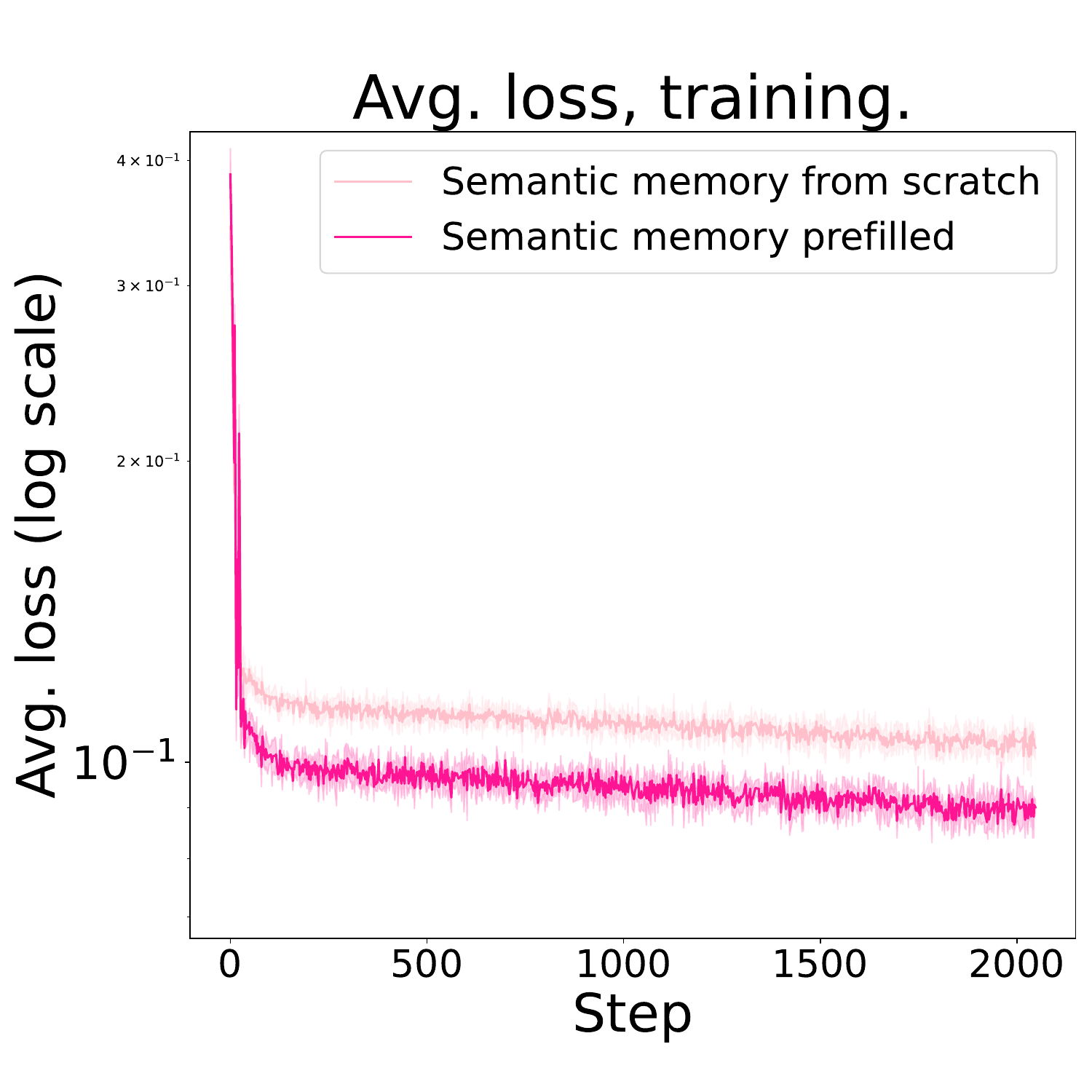}
    \label{fig:train_loss}}
    \hfill  
    \subfigure[Average total rewards per episode, validation.]{
    \includegraphics[width=0.3\textwidth]{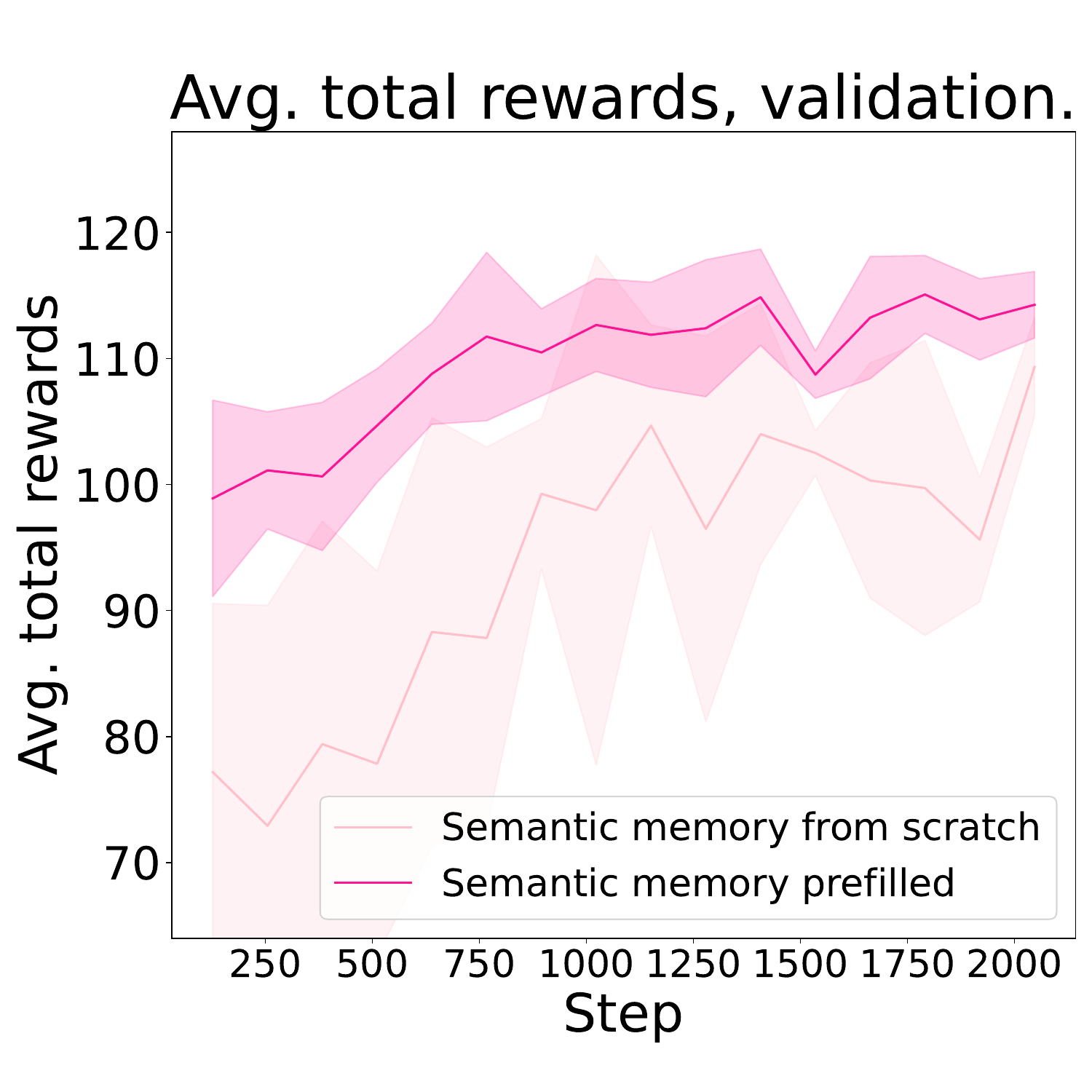} 
    \label{fig:reward_sum_validation}}
    \hfill
    \subfigure[Average total rewards per episode, test.]{
    \includegraphics[width=0.36\textwidth]{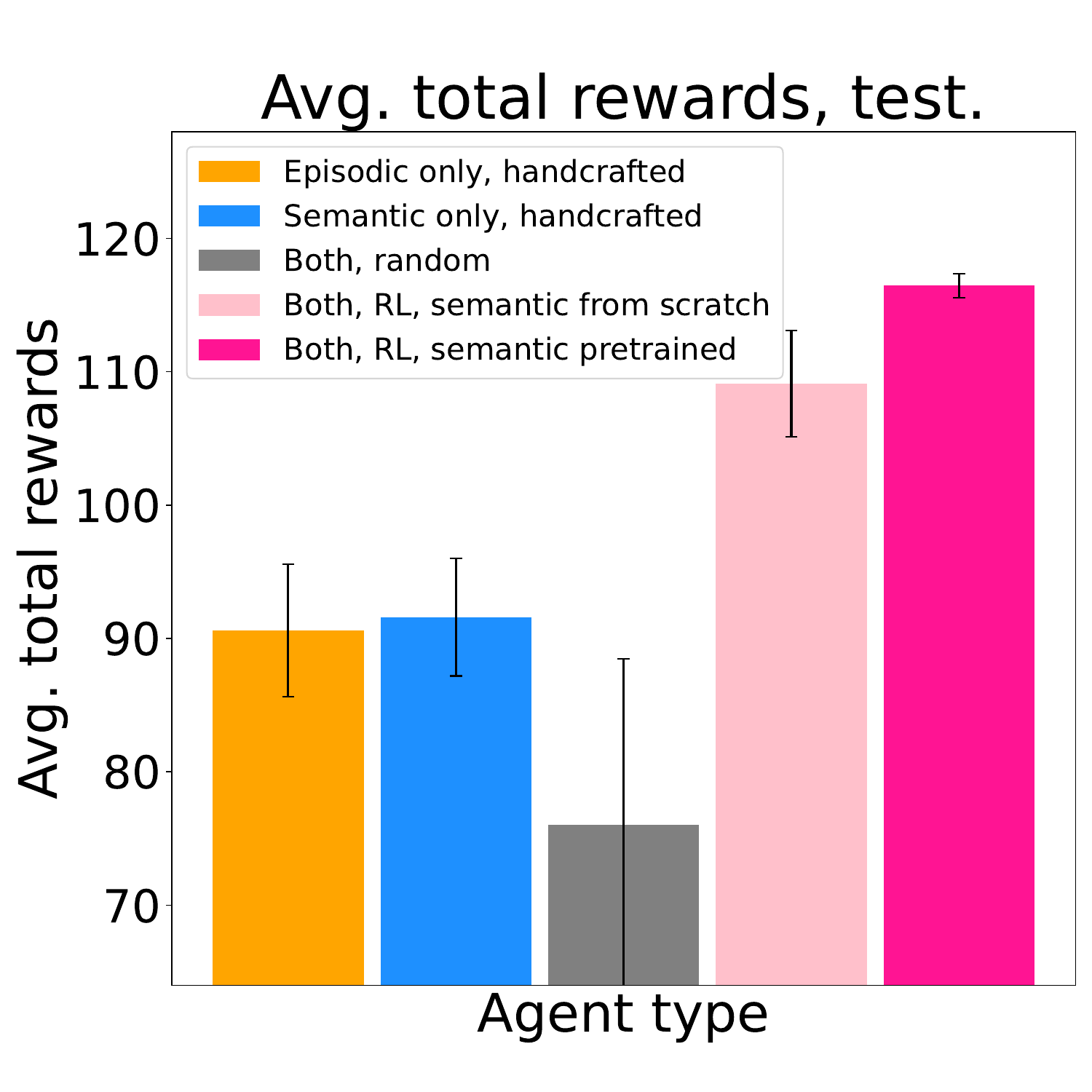} \label{fig:reward_sum_test}}
\caption{Training, validation, and test results of the agents with the memory capacity of 32. 
All of them were run five times with different random seeds. 
The standard deviation is shown together with the average. 
As for training, instead of total rewards per episode, we report loss over time since the varying epsilon value affects the rewards. 
The maximum total rewards per episode are 128, since in total of 128 questions are asked per episode.}
\label{fig:training_validation_test}
\end{figure*}

\section{Experimental Setup}
\label{sec:experimentalsetup}

\subsection{The Environment and Agent Parameters}
\label{sec:room_env_parameters}

In our experimental environment, we set $N_{humans} = 64$, $N_{objects} = 16$, and $N_{object\_locations} = 28$. We determined these values to be sufficiently challenging. As for $p_{commonsense}$, we set it as $0.5$, to avoid bias by giving equal weights to commonsense and non-commonsense probabilities, and thus avoiding bias.

The environment terminates after 128 steps, meaning that the agent will make 128 observations and be asked a total of 128 questions.
This makes the minimum and maximum total episode rewards 0 and 128, respectively. 

We further simplify the agent's observation, $(h, r, t, timestamp)$ because we only use \texttt{AtLocation} for the relation $r$. 
Since this would be the same for all the memory systems, we set the relation embedding to 0. 

The agent observes the humans in a fixed order (e.g., Alice → Bob → Alice → Bob), this ensures that it observes all of them roughly equally often. 
The questions about object locations that the agent has observed are sampled uniformly at random.
This means that an agent with infinite and perfect memory would be able to always answer these questions perfectly.

We train RL agents with different memory capacities. 
The short-term memory capacity is fixed at 1, while we try episodic and semantic capacity from 1, 2, 4, 8, 16, and to 32. For example, the agent with the biggest memory capacity is
$|\bm{M_{o}}|=1$, $|\bm{M_{e}}|=32$, and $|\bm{M_{s}}|=32$, where $\bm{M_{o}}$, $\bm{M_{e}}$, and $\bm{M_{s}}$ stand for the short-term, episodic, and semantic memory systems, respectively. 
In the experiments, we give an equal capacity for the episodic and semantic memory systems. 
We will refer to the agent's total memory capacity as $|\bm{M_{e}}| + |\bm{M_{s}}|$.

We train two variants of our agents: one variant starts with an empty semantic memory, while the other variant starts with general world knowledge from ConceptNet in its semantic memory system. 
The agent with its semantic memory system prefilled can be considered as a ``pretrained'' agent.
With these variants, we want to investigate whether agents that already know about the world would learn differently from those that start from scratch.

At test time, the policy of a trained agent given a state is to choose the action that has the highest Q-value (i.e., the one the agent considers to be most ``valuable''). 
We compare our two trained agents with the three baselines.
The first one always moves the short-term memory contents to episodic memory.
The second, on the other hand, always chooses to move to semantic memory.
The third one is a random baseline that chooses one of the three actions (i.e., including dropping the memory) uniform at random. 

\subsection{The Deep Q-learning and Other Training Parameters}
\label{sec:learning_parameters}

Our training objective follows that of the original deep Q-learning: to minimize the temporal difference loss (See Equation~\ref{eq:loss}). Some of the few differences are that we use the Huber loss~\cite{10.1214/aoms/1177703732}, instead of the squared error loss, and that we use Adam~\cite{kingma2014method} as an optimizer. 
We found these to achieve slightly better performance.

\begin{equation}
\begin{split}
&L(\theta) = \HuberLoss\Bigl(Q_{\text{target}}, Q\bigl(s,a;\theta)\Bigl)\\
&Q_{\text{target}} = r + \gamma \max_{a'} Q(s', a'; \theta^{\text{target}})\\
\end{split}
\label{eq:loss}
\end{equation}
where $s$, $a$, $s'$, $a'$, $r$, and $\gamma$ denote state, action, next state, next action, reward, and discount factor, respectively. $\theta^{target}$ is the Q-network parameters from $k$ steps before.

We use PyTorch Lighting as our deep learning framework~\cite{falcon2019pytorch}. See Appendix~\ref{sec:hyperparameters} for the complete list of the hyperparameters.

\begin{figure*}[bth]
    \centering
    \includegraphics[width=\textwidth]{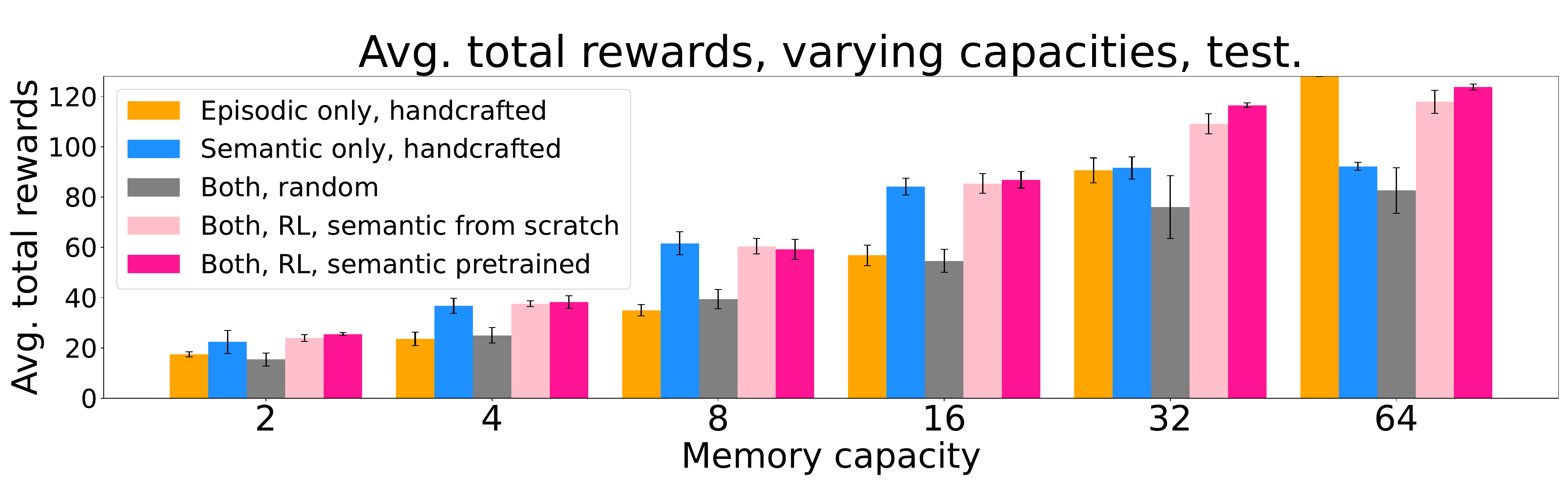}
    \caption{Average total test rewards of the agents with different memory capacities. 
All of them were run five times with different random seeds. 
The error bars indicate the standard deviation. 
The maximum total rewards per episode are 128, since in total of 128 questions are asked per episode. 
The detailed numbers can be found in Appendix~\ref{sec:total_rewards_numbers}.}
    \label{fig:reward_sum_test_all}
\end{figure*}

\section{Results}
\label{sec:results}

\subsection{Deep Q-learning Performance by Agent}
\label{sec:results_training_validation_test}

We know that the agent observes 16 different kinds of objects in the environment. 
Therefore, we first look at the agents with a memory capacity of 32. 
Ideally, these agents will use their semantic memory capacity of 16 to learn the general world knowledge and the other 16 to remember individual-specific episodic memories. The episodic-only and semantic-only baseline agents also have a memory capacity of 32, which is either all episodic or all semantic memory.
To account for the stochasticity of the questions that the environment asks, after each training epoch (i.e., episode) we run it on a validation environment. 
After 16 epochs of training, we choose the model that has the highest total rewards per episode on validation. 
This model is then run again on a test environment to report the final numbers. 
We ran each experiment 5 times. 
Figure~\ref{fig:training_validation_test} shows the results.

We observe that the RL agents with episodic and semantic memory systems outperform all the three baselines. 
It is also interesting to see the performance gap between the two RL agents. 
The one with its semantic memory system pre-filled not only learns faster, but also converges to a better optimum (116 vs. 109 for average total test rewards). This is comparable to transfer learning in neural networks~\cite{5288526}, except that the transferred knowledge in our setting is a symbolic knowledge graph, not neural network parameters.

Next, we investigate the varying size of the memory capacity. 
That is, we want to see how the agents perform if it can store less or more than 32 memory triples, while all other parameters are kept the same. 
Figure~\ref{fig:reward_sum_test_all} shows the results. 
As the memory capacity grows, the episodic-only agent can answer more questions, resulting in higher total rewards. 
If it has high enough memory capacity (e.g., 64), it can indeed answer all 128 questions. 
As for the agent with only semantic memory system, it performs better than the others in the low memory capacity settings. 
This suggests that it is better to focus only on the general world knowledge rather than remembering individuals, when the memory capacity is too limited. 
However, this semantic-only agent's performance plateaus after a certain memory capacity. 
For example, the total test rewards for the memory capacity of 32 and 64 are both about 92. 
This is because 32 general world knowledge triples were already enough to answer the general knowledge related questions, and having more memories has only a minimal impact.

\subsection{Understanding What the Agent Learns}
\label{sec:understanding_what_the_agent_learns}

\begin{figure}[tb]
\centering
\subfigure[The agent with its semantic memory from scratch.]{
\includegraphics[width=\columnwidth]{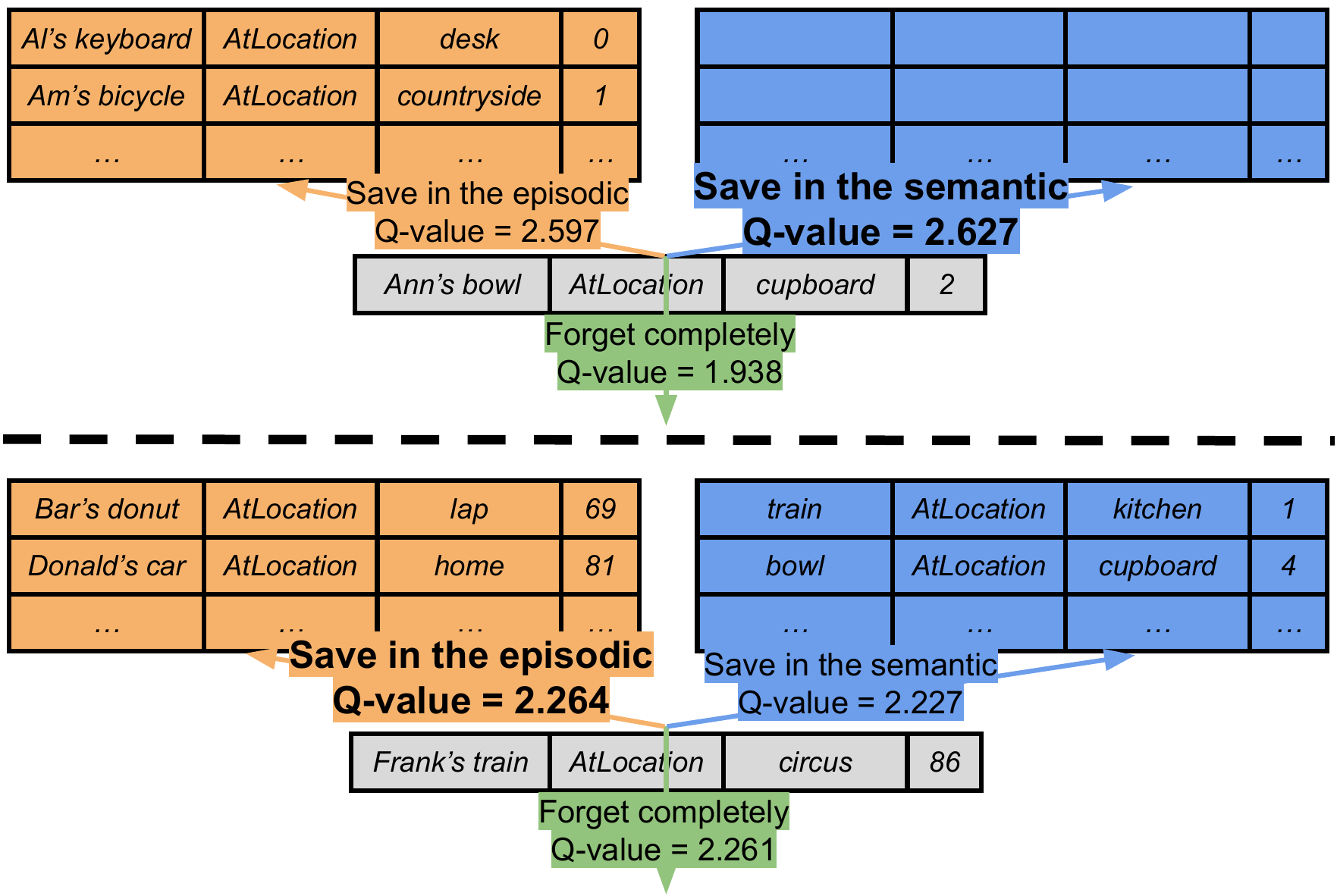}
\label{fig:q_values_scratch_test}
}
\vfill
\subfigure[The agent with its semantic memory prefilled.]{
\includegraphics[width=\columnwidth]{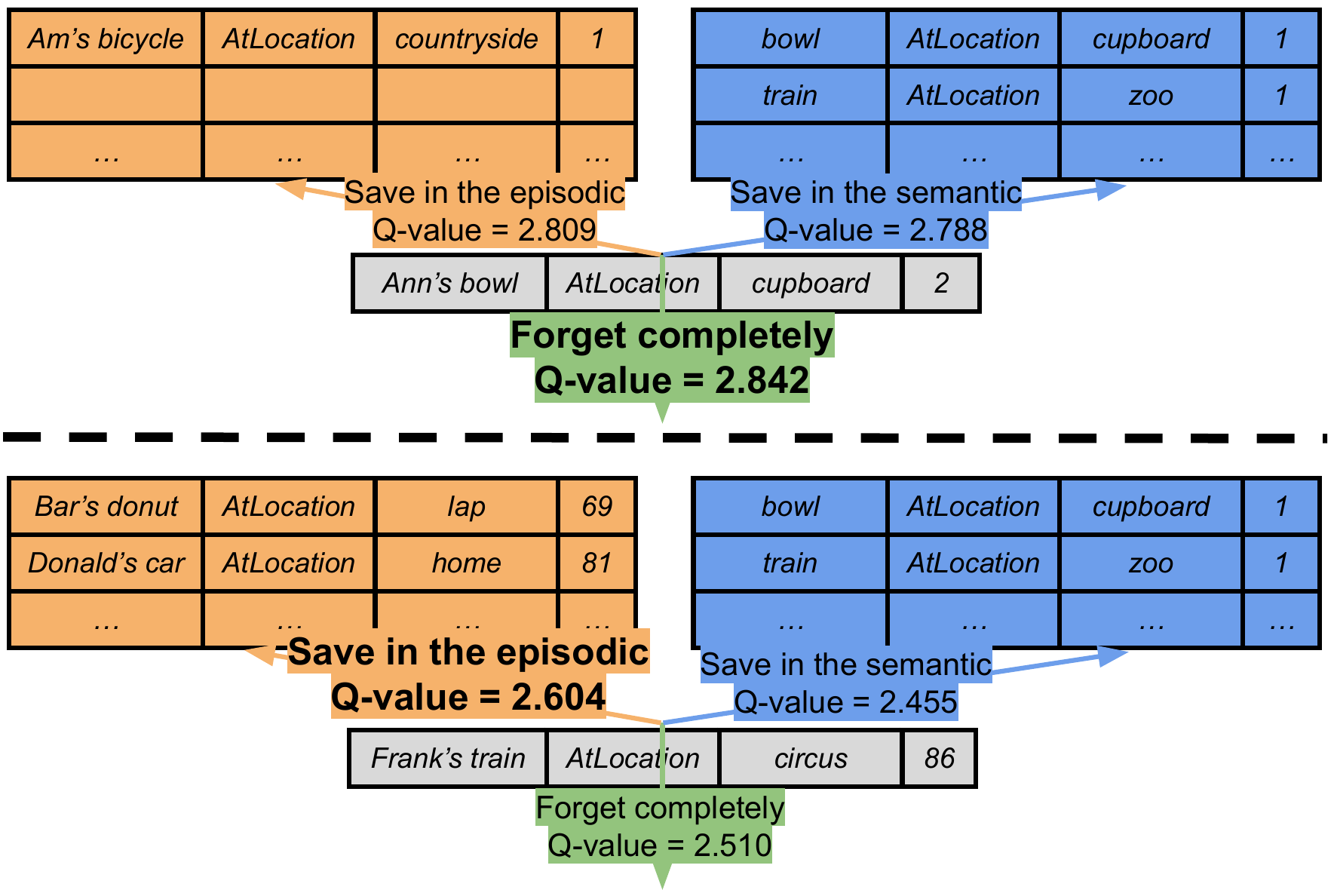}
\label{fig:q_values_pretrained_test}
}
\caption{The Q-values of each of the three possible actions during the test time. 
The short-term, episodic, and semantic memory systems at time step of 2 and 86 are shown together. 
Figure~\ref{fig:q_values_scratch_test} is the agent that starts with its semantic memory from scratch, while Figure~\ref{fig:q_values_pretrained_test} is the one where its semantic memory is prefilled with the general world knowledge.}
\label{fig:q_values_test}
\end{figure}

We display the Q-values during the test time in Figure~\ref{fig:q_values_test}. 
We can see that the semantic-scratch agent starts both of its episodic and semantic memory systems from scratch, while the semantic memory system of the semantic-pretrained agent is already prefilled with the general world knowledge from the beginning. 
There is one same strategy between the two agents. 
Both of them think it is most ``valuable'' to store a not-so-common short-term memory (e.g., \texttt{(Frank's train, AtLocation, circus, 86)}) in the episodic memory system. 
This is understandable, since such an observation is more individual-specific than generalizable. 

But we observe one fundamental difference between them. 
Whenever the semantic-scratch agent observes something that is generalizable (e.g., \texttt{(Ann's bowl, AtLocation, cupboard, 2)}), it stores it in the semantic memory system. 
However, the semantic-prefilled agent never decides to store any short-term memory in the semantic memory system. 
It always forgets generalizable short-term memories completely. 
This makes its semantic memory system untouched. 
As it is filled in the beginning, it stays the same until the end. 
What we observe is that it retains the general world knowledge it was given.
Then, it either stores short-term memory in the episodic one or forgets it completely. 
In our experiments, this strategy indeed results in receiving higher rewards, since we prefilled its semantic memory system with the perfect world knowledge. 
As for the semantic-scratch agent, although it manages to learn most of the general world knowledge, it also makes mistakes. 
For example, it thinks that \texttt{(train, AtLocation, kitchen)} is generalizable, while actually it should have been \texttt{(train, AtLocation, zoo)}. This kind of mistakes results in answering some questions incorrectly, leading to lower total rewards per episode.

\section{Related Work}
\label{sec:related}

Our work stands somewhere between cognitive architecture and machine learning with memory systems.
Cognitive architectures focus on the theory of the human mind, while machine learning with memory systems try to take advantage of memory systems to solve problems such as partial observability and catastrophic forgetting.
Here we list the works that are the most relevant to us.

ACT-R~\cite{10.1093/acprof:oso/9780195324259.001.0001} and Soar~\cite{10.5555/2222503} study the human memory systems comprehensively and provide opportunities for AI agents. 
However, they lack in computational methods, such as training an RL agent with memory systems. 

Gorski and Laird~\cite{GORSKI2011144} have applied RL within the Soar cognitive architecture. While our work focuses on learning a memory storage policy, they have focused on learning episodic memory retrieval. Li~\cite{justinliacs2020} has extended their work by applying it to bigger data.

The Tensor Brain~\cite{DBLP:journals/corr/abs-2109-13392} focuses on the computational theory of an agent's perception, episodic memory, semantic memory. 
They introduced the model named bilayer tensor network (BTN) where perception, episodic memory, and semantic memory are realized. 
They used the VRD dataset~\cite{lu2016visual} to carry out their experiments. Although this dataset includes image modality, it is similar to ours in that it is a form of triples that can form a knowledge graph.
In their experiments, they used supervised learning to predict the relationships between the entities.

Hemmer and Steyvers~\cite{Hemmer2009IntegratingEA} have studied how episodic and semantic memory systems play a role in recall of objects, but their experiments were human-based empirical results, which does not scale as well as our computational method.

Memory-based RL~\cite{10.1109/IROS51168.2021.9636140, 10.5555/3304889.3304998} adds memory components to their classical RL problems to address the partial observability problem and to achieve sample efficiency, respectively. 
Episodic Memory Deep Q-Networks~\cite{10.5555/3304889.3304998} uses the term ``episodic memory'' that is used to regularize their deep Q-network. They use the term episodic memory to refer to the state-action pairs that the agent has taken along the training episode.

Episodic Memory Reader~\cite{han-etal-2019-episodic} trains an RL agent that learns what to remember in their memory system, and they also use question-answering to evaluate their method. 
Their work differs from ours in that they have only one memory system and that it is not made of a knowledge graph of triples, but rather numeric embeddings, which are harder to interpret.

When a deep neural network trained to solve task A is retrained to solve task B, it often forgets what it has learned previously, which is called \textit{catastrophic forgetting}.
Continual learning (also known as  lifelong learning) is concerned with training a neural network on a sequence of tasks without forgetting them. Continual learning with an episodic memory system~\cite{NEURIPS2020_0b5e29aa, 10.5555/3295222.3295393} addresses this issue by storing some previous task examples in their episodic memory system. 
Adjusting the neural network weights for the current task also accounts for the previous task examples, to avoid forgetting them. 

Similar to our work, CLS-ER~\cite{arani2022learning} also takes advantage of the dual memory system (i.e., episodic and semantic) for their continual learning problem. Their episodic memory system stores data samples, while the semantic stores previous model weights.

\section{Conclusions}
\label{sec:conclusions}

Current machine learning agents with memory systems cannot answer all types of questions. 
They are especially weak at answering personal questions. 
To tackle this problem, we investigated an agent with human-like memory systems, inspired by the cognitive science theory. 
Our agent has short-term, episodic, and semantic memory systems, each of which is modeled with a knowledge graph. 

To experiment with this agent, we have designed and released our own environment, ``the Room'', compatible with OpenAI Gym. 
We successfully trained an agent that has interpretable memory systems, with a deep Q-learning based reinforcement learning algorithm. 
This experiment confirms that a machine with human-like memory systems outperforms the ones without this structure, when both general world knowledge and individual-specific questions are asked.

In the future, we would like to make our environment not only more complex (e.g., include more relations in its triples), but also multimodal (e.g., images). 
The questions given by the environment can extend to natural language with ambiguity, not just triples. 
We would also like to add other kinds of human-like memory systems to our agent (e.g., working memory, implicit memory, etc.).
It would also be interesting to include real humans in the loop, so that the agent can ask them a question when it is uncertain of the actions that it can take.

\appendix

\section{Hyperparameters}
\label{sec:hyperparameters}

\begin{table}[tb]
\resizebox{\columnwidth}{!}{\begin{tabular}{l|l}
\hline
Maximum number of epochs (episodes) & 16                       \\ \hline
Batch size                          & 1024                     \\ \hline
Epoch length                        & 1024$\times$128          \\ \hline
Replay size                         & 1024$\times$128          \\ \hline
Warm start size                     & 1024$\times$128          \\ \hline
Epsilon start value                 & 1.0                      \\ \hline
Epsilon end value                   & 0                        \\ \hline
Epsilon last step                   & 128$\times$16            \\ \hline
Gamma (discount factor)             & 0.65                     \\ \hline
Learning rate                       & 0.001                    \\ \hline
DQN sync rate                       & 10                       \\ \hline
Loss function                       & Huber                    \\ \hline
Optimizer                           & Adam                     \\ \hline
Neural network architecture         & LSTM + MLP               \\ \hline
Embedding dimension                 & 32                       \\ \hline
LSTM hidden size dimension          & 64                       \\ \hline
Number of LSTM layers               & 2                        \\ \hline
Processing unit                     & 1 CPU                    \\ \hline
Floating point precision            & 32 bits                  \\ \hline
Number of iterations for evaluation & 10                       \\ \hline
Number of runs                      & 5                        \\ \hline
\end{tabular}}
\caption{The full list of the hyperparameters used.}
\label{tab:hyperparameters}
\end{table}

Table~\ref{tab:hyperparameters} lists all the hyperparameters used. 
Since every episode terminates after 128 steps (questions) and we are training for 16 epochs, the total number of training steps is 128$\times$16 = 2048. One epoch corresponds to one episode (i.e., 128 steps), where every training step includes computing one temporal difference loss from one batch of 1024 data samples. 
And then Adam~\cite{kingma2014method} makes one weight update.

We set epoch length, replay size, and warm start size as 1024$\times$128. This corresponds to the total number of data samples that the optimizer sees in one epoch. 
The epsilon value of DQN decays linearly. 
It starts with 1.0 and ends with 0. The time step when it reaches 0 is 128$\times$16, which corresponds to the last (16th) epoch. 
We tried different values for gamma (discount factor), and we found 0.65 to be the best. $Q_{target}$ is synced with $Q_{\theta}$ every 10 steps.

As for the Q-network, the embedding dimension was set to 32, and the LSTM hidden size dimension was set to 64.
The input feature size of the linear layers was also set to 64, so that last hidden state of the LSTM can be directly fed into them.
ReLU~\cite{nair2010rectified} was used for the non-linearity between the linear layers in the MLP. The number of LSTM layers used is 2.

Since our neural network is relatively small (i.e., the number of parameters is 265,000), we just used one CPU (i.e., Intel® Core™ i7-10875H) with 64 GB of RAM, instead of using GPUs or ASICs. 
It took about 3 hours to train one semantic-scratch agent with the memory capacity of 32.

For every validation and test, we ran the model 10 times to get the average of total rewards per episode. 
We did this because the environment is stochastic, and we wanted to get the best estimate. 
This means that the final reported total rewards of the test environment are the average of 10$\times$5 numbers.

\section{Total Test Rewards for All Capacities}
\label{sec:total_rewards_numbers}

\begin{table}[tb]

\resizebox{\columnwidth}{!}{\begin{tabular}{l||l|l|l|l|l}\hline
\backslashbox[50mm]{Agent memory type}{Memory capacity}
& \multicolumn{1}{c|}{\textbf{2}} & \multicolumn{1}{c|}{\textbf{4}}             \\ \hline\hline
\textit{Episodic-only}                 & 17.4 $\pm$ 1.0   & 23.6 $\pm$ 2.7      \\ \hline
\textit{Semantic-only}                 & 22.4 $\pm$ 4.5   & 36.8 $\pm$ 3.0      \\ \hline
\textit{Both, random}                  & 15.4 $\pm$ 2.6   & 25.0 $\pm$ 3.0      \\ \hline
\textit{Both, RL, semantic-scratch}    & 23.9 $\pm$ 1.4   & 37.6 $\pm$ 1.2    \\ \hline
\textit{Both, RL, semantic-pretrained} & 25.5 $\pm$ 0.6   & 38.2 $\pm$ 2.5    \\ \hline
\end{tabular}}

\resizebox{\columnwidth}{!}{\begin{tabular}{l||l|l|l|l|}\hline
\backslashbox[10mm]{}{}
& \multicolumn{1}{c|}{\textbf{8}} & \multicolumn{1}{c|}{\textbf{16}} & \multicolumn{1}{c|}{\textbf{32}} & \multicolumn{1}{c|}{\textbf{64}}             \\ \hline\hline
\textit{} & 35.0 $\pm$ 2.3   & 56.8 $\pm$ 4.1   & 90.6 $\pm$ 5.0    & 128.0 $\pm$ 0.0    \\ \hline
\textit{} & 61.6 $\pm$ 4.6   & 84.2 $\pm$ 3.3   & 91.6 $\pm$ 4.4    & 92.2 $\pm$ 1.6     \\ \hline
\textit{} & 39.4 $\pm$ 3.8   & 54.6 $\pm$ 4.5   & 76.0 $\pm$ 12.5   & 82.6 $\pm$ 9.1     \\ \hline
\textit{} & 60.4 $\pm$ 3.1   & 85.3 $\pm$ 3.9   & 109.1 $\pm$ 4.0   & 117.9 $\pm$ 4.6    \\ \hline
\textit{} & 59.2 $\pm$ 4.0   & 86.8 $\pm$ 2.3   & 116.5 $\pm$ 0.9   & 123.8 $\pm$ 1.2    \\ \hline
\end{tabular}}

\caption{The average total test rewards per episode for all capacities. The standard deviations are shown after $\pm$.}
\label{tab:total_test_rewards}
\end{table}

Table~\ref{tab:total_test_rewards} shows the average total test rewards and their standard deviations per episode for all capacities.

\section*{Acknowledgments}
\label{sec:ack}
This research was (partially) funded by the Hybrid Intelligence Center, a 10-year program
funded by the Dutch Ministry of Education, Culture and Science through the Netherlands Organization
for Scientific Research, \url{https://www.hybrid-intelligence-centre.nl/}.
\bibliography{aaai23}
\end{document}